\documentclass[11 pt]{article}
\usepackage{cite}
\usepackage{amsmath,amssymb,amsfonts}
\usepackage{algorithmic}
\usepackage{graphicx}
\usepackage{textcomp}
\usepackage{xcolor}
\usepackage{subcaption}
\usepackage{authblk}

\usepackage[utf8]{inputenc}  % é
\usepackage{booktabs}       % professional-quality tables
\usepackage{hyperref}

\newcommand{\M}{{\bf M}}
\newcommand{\Id}{{\bf I}}
\renewcommand{\L}{{\bf L}}
\newcommand{\R}{\mathbb{R}}
\newcommand{\Z}{\mathcal{X}}
\newcommand{\X}{\mathcal{X}}
\newcommand{\Y}{\mathcal{Y}}
\newcommand{\LS}{\mathcal{S}}
\newcommand{\Fcal}{{\cal F}}
\newcommand{\Simp}{\textit{Sim}^+}

\newcommand{\Simn}{\textit{Sim}^-}

\newcommand{\Disp}{\textit{Dis}^+}

\newcommand{\Disn}{\textit{Dis}^-}

\newcommand{\D}{\mathcal{D}}

\begin{document}

\title{Metric Learning from Imbalanced Data}

%\author{\IEEEauthorblockN{Léo Gautheron, Amaury Habrard, Emilie Morvant, Marc Sebban} \IEEEauthorblockA{\textit{Univ Lyon, UJM-Saint-Etienne, CNRS, Institut d Optique Graduate School}\\ \textit{Laboratoire Hubert Curien UMR 5516}\\ F-42023, ST-ETIENNE, France \\ leo.gautheron@univ-st-etienne.fr} }

\author{Léo Gautheron, Emilie Morvant, Amaury Habrard and Marc Sebban}
\affil{Laboratoire Hubert Curien UMR 5516,
	Univ Lyon, UJM F-42023,  Saint-Etienne, France. }

\maketitle

\begin{abstract}
A key element of any machine learning algorithm is the use of a function that measures the dis/similarity between data points. 
Given a task, such a function can be optimized with a {\it metric learning} algorithm.
Although this research field has received a lot of attention during the past decade, very few approaches have focused on learning a metric in an imbalanced scenario where the number of positive examples is much smaller than the negatives.
%Since many machine learning algorithms require a distance metric to capture dis/similarities between data points, {\it metric learning} has received much attention during the past decade.
%Surprisingly, very few methods have focused on learning a metric in an imbalanced scenario where the number of positive examples is much smaller than the negatives.
Here, we address this challenging task by designing a new Mahalanobis metric learning algorithm ({\bf IML}) which deals with class imbalance.
The empirical study performed %on a wide range of datasets %wide je trouve ca un peu fort
shows the efficiency of {\bf IML}.
\end{abstract}

\section{Introduction}
Metric learning~\cite{bellet2013survey,bellet2015metric,kulis2013metric} is a subfield of representation learning that consists of designing a pairwise function able to capture the dis/similarity between two data points.
This is a key issue in machine learning where such metrics are at the core of many algorithms, like \mbox{$k$-nearest} neighbors ($k$NN), SVMs, \mbox{$k$-Means}, etc.
To construct a dis/similarity measure suitable for a given task, most metric learning algorithms optimize a loss function which aims at bringing closer examples of the same label while pushing apart examples of different labels. 
In practice, metric learning is usually performed with cannot link/must link constraints---two data points $x$ and $x'$ should be dis/similar~\cite{davis2007information,lu2014neighborhood,weinberger2009distance,xiang2008learning,xing2003distance,zadeh2016geometric}---or relative constraints---a data point $x$ should be more similar to another $x'$ than to a third one $x''$\cite{lee2008rank,schultz2004learning,weinberger2009distance,zheng2011person}.\\

In this paper, we focus on the family of metric learning algorithms that construct a Mahalanobis distance defined as
$$ d_{\M}(x,x')=\sqrt{(x-x')^{\top} \M (x-x')}\,,$$  parameterized by a positive semidefinite matrix $\M$.
Learning such a Mahalanobis distance leads to several nice properties: {\it (i)} $d_{\M}$ is a generalization of the Euclidean distance; {\it (ii)}~it induces a projection such that the distance between two points is equivalent to their Euclidean distance after a linear projection; {\it (iii)}~$\M$ can be low rank implying a projection in a lower dimensional latent space; {\it (iv)} it involves optimization problems that are often convex and thus easy to solve.
The most famous Mahalanobis distance learning algorithms are likely
\textbf{LMNN}~(Large Margin Nearest Neighbor~\cite{weinberger2009distance}) and \textbf{ITML} (Information-Theoretic Metric Learning~\cite{davis2007information}), which are both designed to improve the accuracy of the $k$NN classification rule in the latent space.
The principle of \textbf{LMNN} is the following: for each training example, its  $k$ nearest neighbors of the same class (the {\it target neighbors}) should be closer than examples of other classes (the {\it impostors}).
The algorithm \textbf{ITML} uses a LogDet regularization and minimizes (respectively maximizes) the distance between examples of the same (respectively different) class.
We can also cite another recent Mahalanobis distance learning algorithm called \textbf{GMML}~(Geometric Mean Metric Learning~\cite{zadeh2016geometric}) where the metric is computed using a closed form solution of an unconstrained optimization problem involving similar and dissimilar pairs.
In light of these learning procedures, it is worth noticing that the loss functions optimized in \textbf{LMNN}, \textbf{ITML} and \textbf{GMML} (and in most pairwise metric learning methods) tend to favor the majority class as there is no distinction between the constraints involving examples of the majority class and the constraints on the  minority class.
\begin{figure*}[t]
	\centering
	\includegraphics[width=1.00\textwidth]{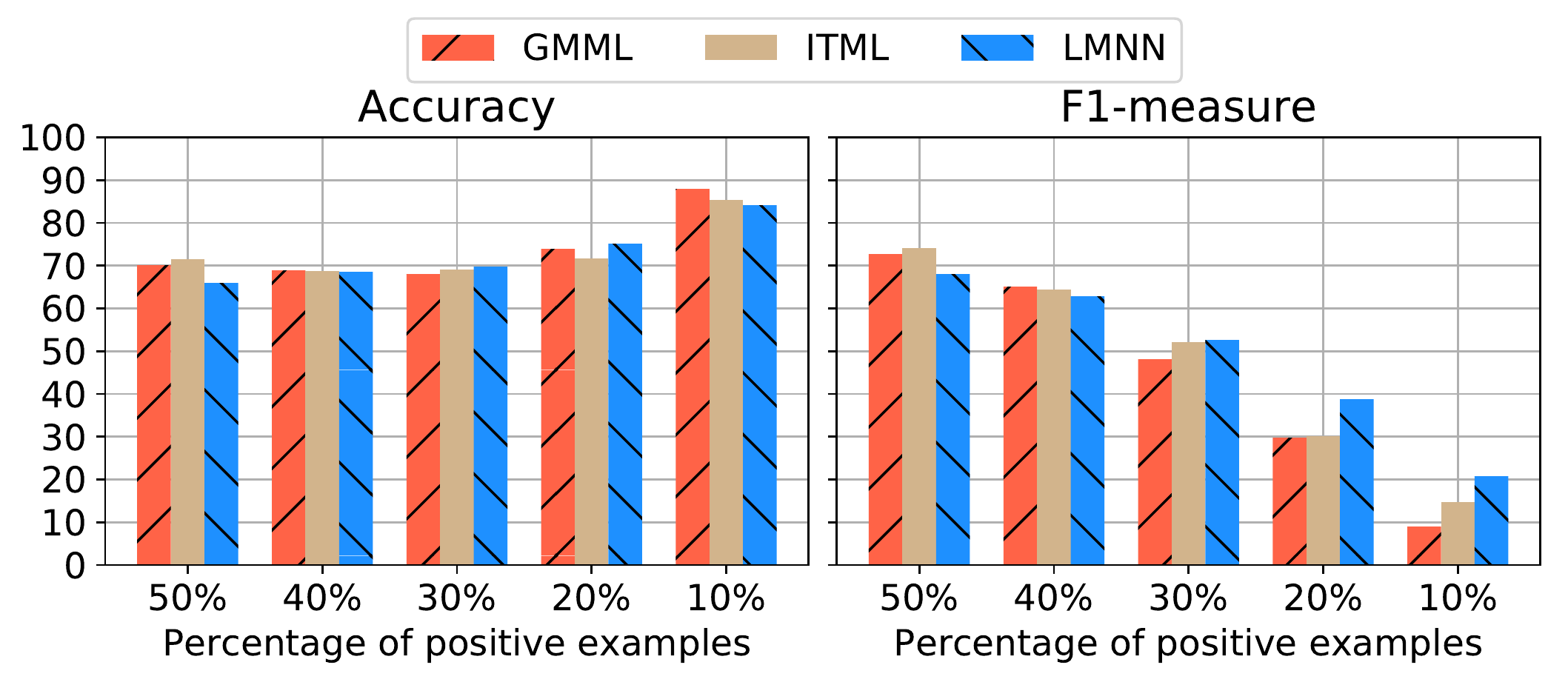}
	\caption{Illustration on the {\sc spectfheart} dataset of the negative impact of classic metric learning algorithms when facing an increasing imbalance in the dataset. On the left, as the proportion of minority examples decreases (the positive class), the $k$NN algorithm with the learned metrics tend to classify all the examples as the majority class, with an accuracy close to $100\%$.
	On the right, using the F1-measure, we note that the classifier actually miss many positives, typically considered as the examples of interest.}\label{fig:increaseImbalance}
\end{figure*}
This strategy is thus not well suited when dealing with imbalanced classes.
An illustration of this phenomenon on the {\sc spectfheart} dataset from the UCI repository is shown in Figure~\ref{fig:increaseImbalance}.
We observe that decreasing the proportion of minority examples tends to generate a metric which classifies (with a $k$NN rule) all the examples as the majority class, thus leading to an accuracy close to $1$.
On the other hand, the F1-measure~\cite{fmesure}, commonly\footnote{The F1-measure is much more adapted to imbalanced scenarios since it does not involve the true negatives but considers both the false positives and the false negatives.} used in imbalanced settings~\cite{chandola2009anomaly,lopez2013insight}, decreases with the proportion of positives, showing that the classifier missed many positives, usually considered as the examples of interest.

This problem of learning from imbalanced data has been widely tackled in the literature~\cite{branco2016survey,he2009learning}.
Classic methods typically make use of over/under-sampling techniques~\cite{drummond2003c4,estabrooks2004multiple,liu2009exploratory,Aggarwal,bauder2018data,pereira2018dealing,ferreira2017improving} or create synthetic samples in the neighborhood of the minority class, {\it e.g.,} using SMOTE-like strategies~\cite{Chawla:2002,chawla2003smoteboost,han2005borderline} or resorting to adversarial techniques~\cite{douzas2018effective}.
However, these methods may lead to over or under-fitting and are often subject to an inability to generate enough diversity, especially in a highly imbalanced scenario.
Other strategies aim at addressing imbalanced situations directly during the learning process. 
They include cost-sensitive methods~\cite{elkan2001foundations,zadrozny2003cost} which require prior knowledge on the miss-classification costs, the optimization of imbalance-aware criteria~\cite{FreryECML2017,mcfee2010metric,vogel2018probabilistic} which are often non convex, or ensemble methods based on bagging and boosting strategies~\cite{galar2012review} that can be computationally expensive.\\

Unlike the state of the art, we propose in this paper to address the problem of learning from imbalanced data by optimizing a metric suited to scenarios where the positive data are very scarce.
As far as we know, very few methods were designed in this setting.
We can cite, Feng {\it et al.}~\cite{feng2018learning} propose to regularize a standard metric learning problem by using the \mbox{KL-divergence} between the classes.
Wang {\it et al.}~\cite{wang2018iterative} use {\bf LMNN} to learn  a classic metric and then perform a sampling on the training data to account the imbalance.
However, as we will see in our experimental study, better performances can be achieved by resorting to a metric dedicated specifically to deal with the imbalance of  the application at hand.
In order to implicitly control the rates of false positives and false negatives, we design a new algorithm, called \textbf{IML} for Imbalanced Metric Learning, which accounts carefully the nature of the pairwise constraints (by decomposing them with respect to the labels involved in the pairs) and gives them an equal weight to account the imbalance.\\

The rest of the paper is organized as follows.
Section~\ref{sec:notations} introduces the notations and the principle of classical Mahalanobis metric learning.
Section~\ref{sec:iml} describes our algorithm {\bf IML} which takes the form of a simple regularized convex problem. 
We perform an experimental study of our approach  in Section~\ref{sec:experiments} before concluding in Section~\ref{sec:conclusion}.

%\section{{\bf IML}: Imbalanced Metric Learning}
%\label{sec:method}
%\subsection{Notations and Setting}
\section{Notations and Setting}
\label{sec:notations}
In this paper, we deal with binary classification tasks where  $\X\subseteq\R^d$ is a \mbox{$d$-dimensional} input space and $\Y=\{-1,+1\}$ is the binary output space. 
We further define $\mathcal{Z}=\X\times \Y$ as the joint space where \mbox{$z=(x,y)\in \mathcal{Z}$} is a labeled example.
In supervised classification, a machine learning algorithm is provided with a learning sample $\LS=\{z_i{=}(x_i,y_i)\}_{i=1}^n$ of $n$ labeled examples {\it i.i.d.} from a fixed yet unknown distribution $\D$ over $\mathcal{Z}$.
We assume that the learning sample is defined as $\LS=\LS^+\!\cup \LS^-$, with $\LS^+$ the set of positive examples and $\LS^-$ the set of negative examples such that the number of positives $n^+=\vert \LS^+\vert$ is smaller than the number of negatives $n^-=\vert \LS^-\vert$ (we say that $+1$ is the minority class and $-1$ the majority one).
We consider a hypothesis space $\Fcal$, such that $\forall f\in\Fcal,\ f:\X\rightarrow\Y$.
Given $\LS$ and $\Fcal$, the final objective of the learner is to find in $\Fcal$ a hypothesis $f$ (called a classifier) 
%: \X\rightarrow\Y$ (from a hypothesis space  $\Fcal$)
which behaves well on \mbox{from $\D$}, meaning that $f$ has to classify correctly  unseen data points.\\

In this work, we aim at constructing a Mahalanobis distance which induces a new space in which a $k$NN classifier will work well on both classes.
The Mahalanobis distance is a type of metric parameterized by a positive semidefinite (PSD) matrix $\M \in \R^{d\times d}$ that can be decomposed as $\M=\L^\top\L$, where $\L\subseteq\R^{r\times d}$ is a projection induced by $\M$ (where $r$ is the rank of $\M$).
A nice property is that the Mahalanobis distance between two points $x$ and $x'$ is equivalent to the Euclidean distance  after having projected $x$ and $x'$ in the $r$-dimensional space, {\it i.e.},
\begin{align*}
	d_{\M}^2(x,x')\ &=\ (x-x')^\top\,\M\,(x-x')\\
	&=\ (\L x-\L x')^\top\,(\L x-\L x')\,.
\end{align*}

%\subsection{Problem Formulation}
Mahalanobis metric learning algorithms \cite{bellet2015metric,cao2016generalization,jin2009regularized} can usually be expressed as follows:
\begin{equation}\label{eq:classic}
	\min_{\M\succeq 0}F(\M) =\frac{1}{n^2}\!\!\sum_{(z,z')\in \LS^2}\!\!\ell(\M,z,z')+\lambda\, Reg(\M),\!
\end{equation}
where one wants to minimize the trade-off between a convex loss $\ell$ over all pairs of examples and a regularization $Reg$ under the PSD constraint $\M\succeq 0$.

The major drawback of this classical formulation is that the loss gives the same importance to any pair of examples  $(z, z')$ whatever the labels $y$ and $y'$.
Intuitively, this is not well suited to imbalanced data where the minority class is the set of examples of interest (think, for example, about anomaly detection~\cite{chandola2009anomaly}).
Note that some metric learning algorithms~\cite{weinberger2009distance,zadeh2016geometric} allow us to weight the role played by the  must-link and cannot-link constraints, but they do not directly take into account the labels of the examples.
%However, the problem still holds because the labels of the examples are not directly taken into account.

To tackle these drawbacks, we propose in the next section {\bf IML}, a metric learning algorithm able to deal with imbalanced data. 

\section{{\bf IML}: Imbalanced Metric Learning}
\label{sec:iml}
Our algorithm is built on the simple idea consisting in decomposing further the sets of must-link and cannot-link constraints based on the two labels involved in the constraints.
Each set can then be weighted differently during the optimization to reduce the negative effect of the imbalance.\\

Starting from \eqref{eq:classic}, we need to define a loss function $\ell$ on which we base our \textbf{IML} algorithm; we have for all $(z,z')\in \Z^2$ and for all $\M \in \R^{d\times d}$ :
%Let us define the loss function $\ell$ associated to our \textbf{IML} algorithm as follows:
\begin{equation}
%\forall (z,z')\in \Z^2,\ \forall \M \in \R^{d\times d},\quad 
	\ell(\M,z,z') = \begin{cases}
		\ell_1(\M,z,z') = \Big[d_{\M}^2(x,x'){-}1\Big]_+& \!\!\!\!\text{if}~y{=}y',\\[2mm]
		\ell_2(\M,z,z') = \Big[1{+}m{-}d_{\M}^2(x,x')\Big]_+& \!\!\!\!\text{if}~y{\neq}y',
	\end{cases}\label{eq:loss}
\end{equation}
where $[a]_+\!=\max(0,a)$ is the Hinge loss and $m\geq 0$ a margin parameter.

\begin{figure}[t]
    \centering
    \includegraphics[width=\columnwidth]{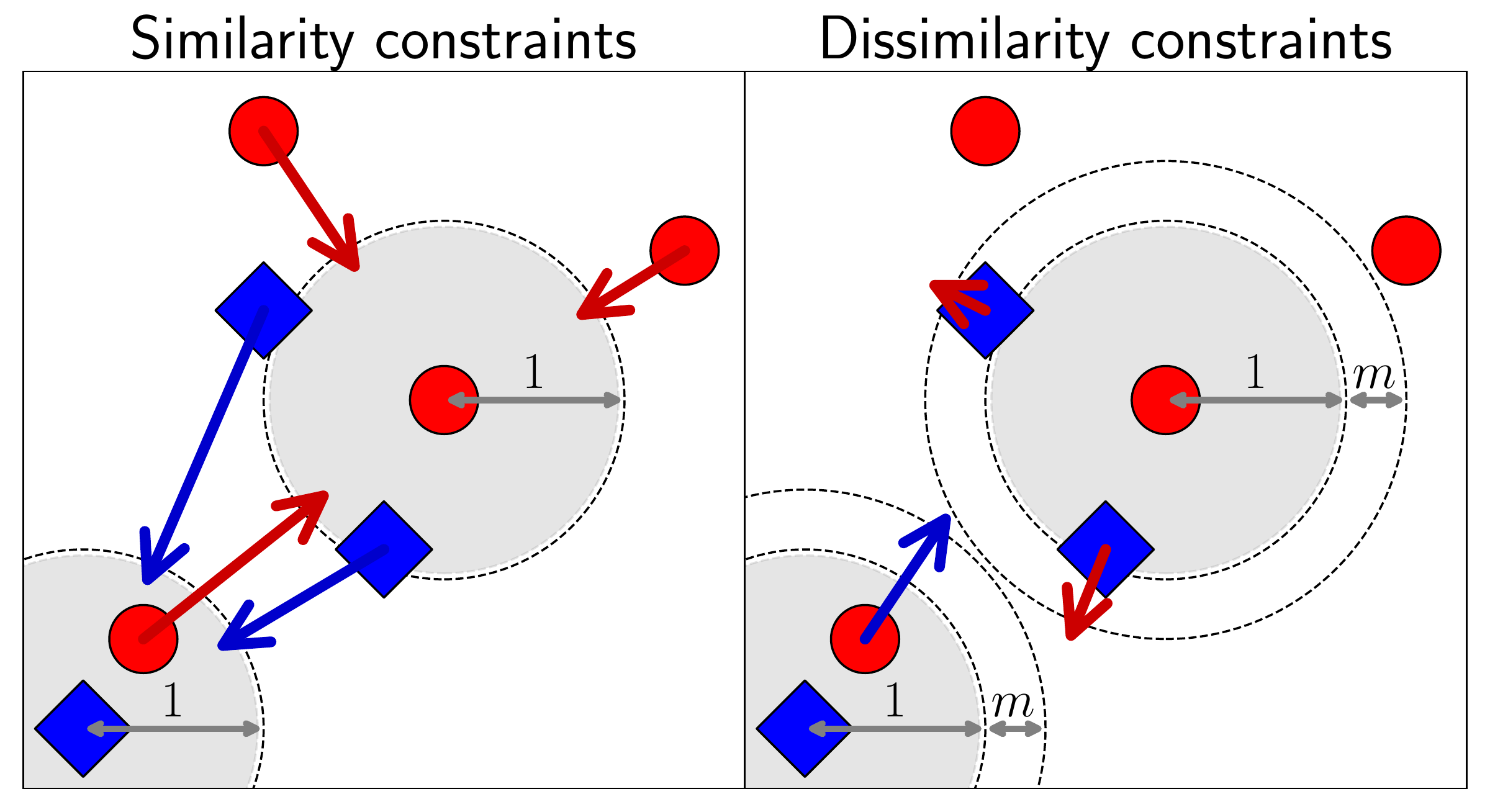}
    \caption{Illustration of the behavior of our loss $\ell$ defined in \eqref{eq:loss}. On the left, the similarity constraints (loss $\ell_1$) aim at bringing examples of the same class at a distance less than $1$. On the right the dissimilarity constraints (loss $\ell_2$) aim at pushing away examples of different classes at a distance larger than $1$ plus a margin $m$.}\label{fig:toyConstraint}
\end{figure}
We illustrate in Figure~\ref{fig:toyConstraint} the behavior of the two losses $\ell_1$ and $\ell_2$.
The idea of $\ell_1$ is to bring examples of the same class at a distance less than $1$ while $\ell_2$ aims to push far away examples of different classes at a distance larger than $1$ plus a \mbox{margin $m$}.

In addition to inserting \eqref{eq:loss} into \eqref{eq:classic}, we need to set the regularization term $Reg(\M)$.
In order to avoid overfitting, we propose to enforce $\M$ to be close to the identity matrix $\Id$ such as  $Reg(\M)=\Vert\M-\Id\Vert^2_F$, with $\Vert\cdot\Vert_F$ the Frobenius norm.
%Moreover, we set the regularization term of \eqref{eq:classic} as $Reg(\M)=\Vert\M-\Id\Vert^2_F$ where $\Id$ is the identity matrix and $\Vert\cdot\Vert_F$ is the Frobenius norm.
%It aims at avoiding overfitting by enforcing $\M$ to be close to the identity matrix $\Id$.
In other words, we aim at learning a Mahalanobis metric which is close to the Euclidean distance while satisfying the best the semantic constraints.

All things considered, our \textbf{IML} algorithm takes the form of the following convex problem:
\begin{align}
	\min_{\M\succeq 0}\ F(\M)\	 =\ &\tfrac{a}{4\vert \Simp \vert}\!\!\!\!\!\! \sum_{(z,z')\in \Simp}\!\!\!\!\!\!\,\ell_1(\M,z,z')\, +\nonumber\\
	&\ \tfrac{a}{4\vert \Simn \vert}\!\!\!\!\!\!\sum_{(z,z')\in \Simn}\!\!\!\!\!\!\!\!\,\ell_1(\M,z,z')\, +\nonumber\\
	&\ \tfrac{(1-a)}{4\vert \Disp \vert}\!\!\!\!\!\!\sum_{(z,z')\in \Disp}\!\!\!\!\!\!\,\ell_2(\M,z,z')\, +\nonumber\\
	&\ \tfrac{(1-a)}{4\vert \Disn \vert}\!\!\!\!\!\!\sum_{(z,z')\in \Disn}\!\!\!\!\!\!\,\ell_2(\M,z,z')\, +\nonumber\\
	&\  \lambda \left\Vert\M-\Id\right\Vert^2_F\,,\label{eq:problem}
\end{align}
where the four sets $\Simp$, $\Simn$, $\Disp$ and $\Disn$ are defined as subsets of $\LS\times\LS$ respectively as: 
\begin{align*}
\Simp\,&\subseteq\,\LS^+\!\times\LS^+\ ,\\
\Simn\,&\subseteq\,\LS^-\!\times\LS^-\ ,\\ \Disp\,&\subseteq\,\LS^+\!\times\LS^-\ ,\\
\mbox{and}\quad \Disn\,&\subseteq\,\LS^-\!\times\LS^+  \ .  
\end{align*}
The parameter $a$ takes values in $[0, 1]$; it controls the trade-off between bringing closer the similar examples and keeping far away the examples of different classes.

The fundamental difference between our formulation and classic metric learning formulations is that we separate in our loss the set of similar pairs into two sets $\Simp$ and $\Simn$, and the set of dissimilar pairs into two sets $\Disp$ and $\Disn$. 
In a classic metric learning formulation, these four sets are all treated equally by giving them a weight of $\tfrac{1}{n^2}$. 
However in the presence of imbalanced data, the number of pairs in $\Simp$ and $\Disp$ which is in $O(n^+)$ is much smaller than in the sets $\Simn$ and $\Disn$ where the number of pairs is in $O(n^-)$. 
Intuitively, in the presence of imbalanced data, the terms in $O(n^+)$ will have a smaller impact on the loss function, thus, we aim at re-weighting these four sets to account the imbalance. 
We adopt a simple strategy consisting in giving a weight to each set that depends on its number of elements. 
We choose to give to the four sets a weight $\frac{1}{4\vert \text{set}\vert}$. 
This strategy allows us to give the same importance to the four terms in the loss function, no matter how imbalanced the data is. 
We will see experimentally that using this re-weighting instead of the weight $\tfrac{1}{n^2}$ greatly increases performances when facing increasingly imbalanced data.

If we look more closely at \eqref{eq:problem}, when all pairs from $\LS\times\LS$ are involved, $\Simp$ and $\Simn$ contain respectively $n^+n^+$ and $n^-n^-$ pairs while $\Disp$ and $\Disn$ contain respectively $n^+n^-$ and $n^-n^+$ pairs.
This means that the pairs in $\Disp$ and $\Disn$ are symmetric and these two sets might be merged.
However, metric learning rarely considers all the possible pairs as it becomes quite inefficient in the presence of a large number of examples.
Possible strategies to select the pairs include a random selection of the pairs~\cite{davis2007information,xiang2008learning,xing2003distance,zadeh2016geometric} or a selection based on the nearest neighbors rule~\cite{lu2014neighborhood,weinberger2009distance}.

For this reason, it might make sense to separate the two sets  $\Disp$ and $\Disn$ and allows us to weight them differently as {\it (i)} they may not consider the same subsets of pairs, and {\it (ii)} may not capture the same geometric information.
Another interpretation of such a decomposition in an imbalanced learning setting is the following: if $z'$ is selected as belonging to the neighborhood of $z$, the minimization of the four terms of \eqref{eq:problem} can be seen as a nice way to implicitly optimize with a $k$NN rule the true positive, false negative, false positive and true negative rates respectively.

Among the two strategies to select the pairs, the selection based on the nearest neighbors is more adapted to an imbalanced scenario as it considers $k$ pairs for each training example both from the majority and minority classes. However, the random strategy just picks at random two examples to create a pair. Then with imbalanced data, it might be possible not to have any similar pair between two minority examples, thus focusing on the majority class. 
We will see experimentally that, as expected, the selection of the pairs based on the nearest neighbors rule performs better.

\section{Experiments}
\label{sec:experiments}
\subsection{Datasets} 
We provide here an empirical study of {\bf IML} on 22 datasets coming mainly from the UCI\footnote{\url{https://archive.ics.uci.edu/ml/datasets.html}} and Keel\footnote{\url{http://sci2s.ugr.es/keel/datasets.php}} repositories except for the `{\sc splice}' dataset which comes from LIBSVM\footnote{\url{https://www.csie.ntu.edu.tw/~cjlin/libsvmtools/datasets/binary.html\#splice}}.
All datasets are normalized such that each feature has a mean of $0$ and a variance of $1$.

\begin{table}[t]
	\caption{\label{tab:datasets} Description of the datasets ($n$: number of examples, $d$: number of features, $c$: number of classes) and the class chosen as positive ($Label$), its cardinality ($n^+$) and its percentage (\%).}
	\centering\small
	\resizebox{1.0\textwidth}{!}{\begin{tabular}{lrrrrrrlrrrrrr}
		\toprule
		Name         &     $n$  &  $d$ &  $c$ &   $Label$   &$n^+$ &    $\%$ & Name         &     $n$  &  $d$ &  $c$ &   $Label$   &$n^+$ &    $\%$\\
		\midrule
		splice       &   3175 & 60 &  2 &        -1 & 1527 & 48.10\% & glass        &    214 & 11 &  6 &         1 &   70 & 32.71\%\\
		sonar        &    208 & 60 &  2 &         R &   97 & 46.64\% & newthyroid   &    215 &  5 &  3 &      2, 3 &   65 & 30.23\%\\
		balance      &    625 &  4 &  3 &         L &  288 & 46.08\% & german       &   1000 & 23 &  2 &         2 &  300 & 30.00\%\\
		australian   &    690 & 14 &  2 &         1 &  307 & 44.49\% & vehicle      &    846 & 18 &  4 &       van &  199 & 23.52\%\\
		heart        &    270 & 13 &  2 &         2 &  120 & 44.44\% & spectfheart  &    267 & 44 &  2 &         0 &   55 & 20.60\%\\
		bupa         &    345 &  6 &  2 &         1 &  145 & 42.03\% & hayes        &    160 &  4 &  3 &         3 &   31 & 19.38\%\\
		spambase     &   4597 & 57 &  2 &         1 & 1812 & 39.42\% & segmentation &   2310 & 19 &  7 &    window &  330 & 14.29\%\\
		wdbc         &    569 & 30 &  2 &         M &  212 & 37.26\% & abalone      &   4177 & 10 & 28 &         8 &  568 & 13.60\%\\
		iono         &    351 & 34 &  2 &         b &  126 & 35.90\% & yeast        &   1484 &  8 & 10 &       ME3 &  163 & 10.98\%\\
		pima         &    768 &  8 &  2 &         1 &  268 & 34.90\% & libras       &    360 & 90 & 15 &         1 &   24 &  6.66\%\\
		wine         &    178 & 13 &  3 &         1 &   59 & 33.15\% & pageblocks   &   5473 & 10 &  5 &   3, 4, 5 &  231 &  4.22\%\\
		\bottomrule
	\end{tabular}}
\end{table}
For the sake of simplicity, we have chosen binary datasets, described in Table \ref{tab:datasets} where the minority class is given by the columns  ``Label''. 
Note that \textbf{IML} can easily be generalized to multiclass problems by learning one metric per class in a standard ``one-versus-all'' strategy, and then applying a majority vote~\cite{scholkopfBV95,vapnik1995nature}.
\subsection{Optimization Details}
Like most Mahalanobis metric learning algorithms, {\bf IML} requires that the learned matrix $\M$ is PSD.
There exist different methods to enforce the PSD constraint~\cite{kulis2013metric}.
A classic solution consists in performing a Projected Gradient Descent where one alternates a gradient descent step and a (costly) projection onto the cone of PSD matrices.
The advantage is that the problem remains convex~\cite{xing2003distance} {\it w.r.t.} $\M$, ensuring that one will attain the optimal solution of the problem by correctly setting the  projection step in the gradient descent.
Another solution~\cite{weinberger2008fast} is based on the fact that if $\M$ is PSD, it can be rewritten as \mbox{$\M=\L^\top\L$}.
Therefore, instead of learning $\M$, one can enforce $\M$ to be PSD in a cheaper way  by directly learning the projection matrix \mbox{$\L\in\R^{r\times d}$} (where $r$ is the rank of $\M$).
This can be done thanks to a gradient descent by computing the gradient of the problem {\it w.r.t.} $\L$ (instead of $\M$).
The implementation\footnote{The code is available here: \url{https://leogautheron.github.io}} we propose is based on this latter approach~\cite{weinberger2008fast} where we make use of the \mbox{L-BFGS-B} algorithm~\cite{zhu1997algorithm} from the SciPy Python library to optimize our problem: it takes as input our initial point (the identity matrix), the optimization problem of \eqref{eq:problem}, and its gradient, then it performs a gradient descent that returns the projection matrix $\L$ minimizing \eqref{eq:problem}.
To prevent us from tuning $r$ and finding the best $r$-dimensional projection space, we set $r=d$ in the experiments. 
Indeed, our main objective here is to learn a robust metric and not to get a sparse solution.

As discussed at the end of Section \ref{sec:iml}, the pairs of examples considered by \textbf{IML} in its four terms are chosen using the nearest neighbors rule. Indeed, we noted experimentally that the algorithms using this strategy (\textbf{LMNN} \cite{weinberger2009distance} and \textbf{IML}) perform better than the ones using a random selection strategy (\textbf{ITML} \cite{davis2007information} and \textbf{GMML} \cite{zadeh2016geometric}).

\subsection{Experimental setup}
All along our experiments, we use a $3$NN classifier (like in \textbf{LMNN}) after projection of the training and test data using the metric learned.
The metrics considered in the comparative study are the Euclidean distance and the ones learned by \textbf{GMML} \cite{zadeh2016geometric}, \textbf{ITML} \cite{davis2007information},
\textbf{LMNN} \cite{weinberger2009distance} and \textbf{IML}.
For each dataset, we generate randomly $20$ stratified splits of $30\%$ training examples  and $70\%$ test data (same class proportions in training and test) and report the mean results over the $20$ splits.
The parameters are tuned by \mbox{$5$-fold} cross-validation on the training set through a grid search using the following parameter ranges:
for \textbf{LMNN}, $\mu \in \{0, 0.05, \dots, 1\}$ ($k$ is fixed to $3$); for \textbf{ITML}, $\gamma \in \{2^{-10:10}\}$; for \textbf{GMML} $t \in \{0, 0.05, \dots, 1\}$; and for \textbf{IML}, $m\in\{1, 10, 100, 1000, 10000\}$, $\lambda \in \{0, 0.01, 0.1, 1, 10\}$ and $a \in \{0, 0.05, \dots, 1\}$ ($k$ is also fixed to $3$).
For \textbf{IML}, we select randomly without replacement $100$ combinations of hyper-parameters and use the one giving the best validation performance. 

\subsection{Experiments}

\begin{table}[t]\centering
	\caption{\label{tab:exp2} Average F1-measure $\pm$ standard deviation over 20 splits obtained using different metric learning algorithms.}
	\resizebox{1.0\columnwidth}{!}{\begin{tabular}{l  c c c c c}
			\toprule
			Dataset     &       Euclidean  &       LMNN       &       ITML       &       GMML       &       IML        \\
\midrule
splice      &  76.3 $\pm$  0.7&  76.3 $\pm$  1.3&  79.7 $\pm$  1.4&  86.5 $\pm$  0.8&  \textbf{87.3} $\pm$  0.6\\
sonar       &  69.2 $\pm$  5.3&  69.2 $\pm$  5.0&  70.6 $\pm$  5.9&  70.6 $\pm$  6.5&  \textbf{72.8} $\pm$  5.0\\
balance     &  87.4 $\pm$  1.8&  90.4 $\pm$  1.3&  93.0 $\pm$  1.4&  89.8 $\pm$  1.3&  \textbf{93.1} $\pm$  2.2\\
australian  &  79.9 $\pm$  1.7&  80.9 $\pm$  2.5&  \textbf{82.0} $\pm$  1.9&  81.7 $\pm$  2.0&  82.0 $\pm$  2.0\\
heart       &  76.8 $\pm$  2.1&  76.9 $\pm$  3.6&  76.8 $\pm$  2.9&  74.8 $\pm$  3.2&  \textbf{77.0} $\pm$  3.0\\
bupa        &  49.8 $\pm$  4.4&  \textbf{52.0} $\pm$  5.3&  51.3 $\pm$  4.8&  50.1 $\pm$  5.0&  51.2 $\pm$  5.8\\
spambase    &  85.3 $\pm$  0.9&  86.8 $\pm$  0.8&  87.8 $\pm$  1.0&  88.4 $\pm$  0.8&  \textbf{89.3} $\pm$  0.6\\
wdbc        &  94.2 $\pm$  1.3&  94.4 $\pm$  1.3&  94.3 $\pm$  1.1&  93.5 $\pm$  1.7&  \textbf{94.9} $\pm$  1.3\\
iono        &  67.8 $\pm$  6.7&  72.0 $\pm$  5.4&  73.4 $\pm$  5.4&  70.8 $\pm$  3.9&  \textbf{74.1} $\pm$  4.0\\
pima        &  56.2 $\pm$  1.9&  56.7 $\pm$  3.0&  57.5 $\pm$  3.0&  55.9 $\pm$  3.3&  \textbf{57.6} $\pm$  3.0\\
wine        &  94.9 $\pm$  2.2&  95.5 $\pm$  2.9&  \textbf{96.3} $\pm$  3.3&  96.0 $\pm$  2.9&  95.5 $\pm$  2.4\\
glass       &  66.0 $\pm$  3.4&  \textbf{67.2} $\pm$  3.5&  62.6 $\pm$  5.2&  63.6 $\pm$  5.2&  66.1 $\pm$  3.7\\
newthyroid  &  83.4 $\pm$  4.2&  \textbf{90.6} $\pm$  2.6&  89.8 $\pm$  5.2&  88.1 $\pm$  5.2&  90.5 $\pm$  4.7\\
german      &  35.3 $\pm$  2.8&  37.1 $\pm$  3.9&  37.4 $\pm$  3.3&  37.3 $\pm$  3.9&  \textbf{38.0} $\pm$  3.5\\
vehicle     &  80.5 $\pm$  2.4&  90.1 $\pm$  1.7&  90.2 $\pm$  2.4&  \textbf{92.6} $\pm$  1.0&  91.4 $\pm$  1.9\\
spectfheart &  34.8 $\pm$ 12.3&  29.3 $\pm$ 11.6&  34.4 $\pm$  7.9&  39.1 $\pm$  8.4&  \textbf{45.3} $\pm$  6.4\\
hayes       &  44.9 $\pm$ 13.2&  52.7 $\pm$ 10.8&  55.4 $\pm$  8.7&  \textbf{57.2} $\pm$ 12.5&  56.0 $\pm$ 11.4\\
segmentation&  81.8 $\pm$  2.4&  80.8 $\pm$  3.1&  79.6 $\pm$  3.0&  85.3 $\pm$  2.1&  \textbf{86.6} $\pm$  2.0\\
abalone     &  22.6 $\pm$  2.1&  21.7 $\pm$  1.7&  21.2 $\pm$  3.0&  22.1 $\pm$  2.1&  \textbf{22.6} $\pm$  1.6\\
yeast       &  73.2 $\pm$  2.3&  73.6 $\pm$  2.5&  74.2 $\pm$  3.1&  74.9 $\pm$  2.8&  \textbf{74.9} $\pm$  3.0\\
libras      &  48.4 $\pm$ 15.1&  56.1 $\pm$ 16.3&  65.5 $\pm$ 15.3&  \textbf{68.3} $\pm$ 12.2&  66.8 $\pm$ 13.4\\
pageblocks  &  71.9 $\pm$  3.0&  \textbf{73.7} $\pm$  2.9&  69.7 $\pm$  5.1&  71.8 $\pm$  3.2&  72.4 $\pm$  3.0\\
\midrule
Mean        &  67.3 $\pm$  4.2&  69.3 $\pm$  4.2&  70.1 $\pm$  4.3&  70.8 $\pm$  4.1&  \textbf{72.1} $\pm$  3.8\\
\midrule
Average Rank & 4.30 & 3.26 & 3.00 & 2.96 & 1.48\\
\bottomrule
	\end{tabular}}
\end{table}

\paragraph{First experiment---without data pre-processing}
We start by applying the experimental setup described above and we report the results in Table~\ref{tab:exp2}.
On average, the F1-measure of $72.1\%$  obtained by {\bf IML} is the best in comparison to  $70.8\%$ for {\bf LMNN}, $70.1\%$ for {\bf ITML}, $69.3\%$ for {\bf GMML} and $67.3\%$ for the Euclidean distance. 
Overall, {\bf IML} shows also the best average rank of $1.48$.
We note that \textbf{IML} generally gives better performances on the datasets considered no matter how much they are balanced or not.
This means that our re-weighting scheme of the pairs can not only improve the performances in an imbalanced setting but can be also competitive in more classic scenarios. 

\paragraph{Second experiment---with data pre-processing}
To address imbalanced data issues, classic machine learning algorithms typically resort to over/under-sampling techniques~\cite{Aggarwal} or create synthetic samples in the neighborhood of the minority class---for example by using SMOTE-like strategies~\cite{Chawla:2002}.
We now aim at studying the behavior of those methods when used as a pre-process of the metric learning procedures.
We consider the results of Table~\ref{tab:exp2} as baselines. 
We compare them to the performances obtained after performing prior to metric learning an over-sampling using SMOTE and a Random Under Sampling (RUS) strategy of the negative data.
We use the implementations of these methods from the Python library {\it imbalanced-learn}~\cite{lemaitre2017imbalanced}.

\begin{table*}[t]\centering
\caption{Average F1-measure $\pm$ standard deviation over 20 splits using different metric learning algorithms, but with a pre-processing of the data.}
\begin{subtable}{0.49\textwidth}
\centering
	\caption{\label{tab:exp3_1}
	With a smote pre-processing \cite{Chawla:2002} until $n^+=n^-$.
	}
	\resizebox{1.0\textwidth}{!}{
		\begin{tabular}{l  c c c c c}
			\toprule
			Dataset     &       Euclidean  &       LMNN       &       ITML       &       GMML       &       IML        \\
\midrule
splice      &  74.9 $\pm$  0.9&  76.4 $\pm$  1.3&  79.6 $\pm$  1.2&  86.3 $\pm$  0.8&  \textbf{87.3} $\pm$  0.8\\
sonar       &  72.6 $\pm$  4.2&  71.2 $\pm$  4.2&  72.6 $\pm$  4.5&  73.0 $\pm$  6.4&  \textbf{75.4} $\pm$  4.2\\
balance     &  87.4 $\pm$  1.9&  89.7 $\pm$  1.9&  92.1 $\pm$  1.4&  89.6 $\pm$  1.5&  \textbf{92.4} $\pm$  2.3\\
australian  &  80.3 $\pm$  1.6&  81.1 $\pm$  1.8&  \textbf{82.4} $\pm$  1.5&  80.7 $\pm$  3.2&  81.8 $\pm$  2.0\\
heart       &  77.3 $\pm$  2.0&  \textbf{77.4} $\pm$  3.7&  75.7 $\pm$  4.1&  75.0 $\pm$  2.6&  77.3 $\pm$  2.0\\
bupa        &  54.1 $\pm$  3.1&  \textbf{55.9} $\pm$  4.1&  53.9 $\pm$  3.7&  55.4 $\pm$  3.4&  55.2 $\pm$  3.2\\
spambase    &  85.9 $\pm$  0.7&  87.2 $\pm$  0.8&  87.4 $\pm$  0.9&  88.5 $\pm$  0.6&  \textbf{89.4} $\pm$  0.7\\
wdbc        &  93.4 $\pm$  1.3&  93.6 $\pm$  1.5&  94.0 $\pm$  1.4&  93.5 $\pm$  2.2&  \textbf{94.4} $\pm$  1.5\\
iono        &  78.4 $\pm$  2.6&  78.3 $\pm$  3.9&  77.5 $\pm$  3.8&  77.7 $\pm$  3.7&  \textbf{79.4} $\pm$  3.7\\
pima        &  60.1 $\pm$  2.6&  60.3 $\pm$  2.8&  60.8 $\pm$  2.2&  60.1 $\pm$  2.1&  \textbf{61.3} $\pm$  1.8\\
wine        &  92.7 $\pm$  2.8&  94.7 $\pm$  3.0&  \textbf{96.3} $\pm$  2.6&  95.3 $\pm$  3.0&  96.2 $\pm$  3.1\\
glass       &  66.6 $\pm$  2.9&  \textbf{67.2} $\pm$  3.9&  64.6 $\pm$  3.2&  66.1 $\pm$  4.0&  64.8 $\pm$  6.4\\
newthyroid  &  87.6 $\pm$  3.5&  89.6 $\pm$  4.1&  \textbf{91.6} $\pm$  3.2&  88.7 $\pm$  4.0&  91.0 $\pm$  3.3\\
german      &  46.3 $\pm$  2.2&  46.0 $\pm$  2.3&  46.4 $\pm$  1.8&  45.4 $\pm$  3.5&  \textbf{46.6} $\pm$  1.9\\
vehicle     &  80.6 $\pm$  2.1&  89.5 $\pm$  2.1&  89.9 $\pm$  3.1&  \textbf{92.0} $\pm$  1.6&  91.0 $\pm$  2.3\\
spectfheart &  47.4 $\pm$  2.3&  49.1 $\pm$  4.4&  46.7 $\pm$  6.9&  41.9 $\pm$  8.5&  \textbf{49.6} $\pm$  6.0\\
hayes       &  68.0 $\pm$  6.8&  \textbf{69.6} $\pm$  7.4&  67.8 $\pm$  7.8&  64.4 $\pm$  7.6&  67.7 $\pm$  8.7\\
segmentation&  82.0 $\pm$  1.9&  81.5 $\pm$  1.8&  81.6 $\pm$  2.2&  83.8 $\pm$  2.9&  \textbf{84.5} $\pm$  3.0\\
abalone     &  \textbf{32.3} $\pm$  0.7&  31.7 $\pm$  1.1&  31.9 $\pm$  0.8&  31.4 $\pm$  1.7&  31.9 $\pm$  1.1\\
yeast       &  65.9 $\pm$  2.9&  68.3 $\pm$  2.6&  70.4 $\pm$  2.8&  67.1 $\pm$  3.7&  \textbf{70.5} $\pm$  2.9\\
libras      &  68.3 $\pm$  8.1&  69.3 $\pm$ 10.9&  69.7 $\pm$ 13.8&  76.7 $\pm$  8.5&  \textbf{77.9} $\pm$ 12.2\\
pageblocks  &  62.0 $\pm$  2.9&  61.6 $\pm$  3.5&  55.5 $\pm$  4.0&  61.5 $\pm$  4.1&  \textbf{62.5} $\pm$  3.7\\
\midrule
Mean        &  71.1 $\pm$  2.7&  72.2 $\pm$  3.3&  72.2 $\pm$  3.5&  72.5 $\pm$  3.6&  \textbf{74.0} $\pm$  3.5\\
\midrule
Average Rank & 3.78 & 3.09 & 3.09 & 3.43 & 1.61\\
\bottomrule
	\end{tabular}}
	\end{subtable}
	\begin{subtable}{0.49\textwidth}
	\centering
	\caption{\label{tab:exp3_2}
		With a Random Under Sampling of the negative examples until $n^-=n^+$.}
	\resizebox{1.0\textwidth}{!}{
		\begin{tabular}{l  c c c c c}
			\toprule
			Dataset     &       Euclidean  &       LMNN       &       ITML       &       GMML       &       IML        \\
\midrule
splice      &  75.9 $\pm$  0.7&  76.2 $\pm$  1.2&  79.5 $\pm$  1.6&  86.5 $\pm$  0.6&  \textbf{87.4} $\pm$  0.7\\
sonar       &  70.4 $\pm$  5.2&  69.9 $\pm$  5.2&  70.2 $\pm$  5.7&  73.1 $\pm$  6.3&  \textbf{73.2} $\pm$  4.5\\
balance     &  87.5 $\pm$  1.5&  90.1 $\pm$  1.8&  \textbf{92.8} $\pm$  1.5&  90.1 $\pm$  1.3&  92.8 $\pm$  1.9\\
australian  &  80.4 $\pm$  1.7&  81.5 $\pm$  2.3&  82.2 $\pm$  1.6&  81.7 $\pm$  2.2&  \textbf{82.3} $\pm$  2.3\\
heart       &  \textbf{77.4} $\pm$  2.0&  77.3 $\pm$  1.9&  76.6 $\pm$  2.5&  75.8 $\pm$  3.3&  77.3 $\pm$  2.6\\
bupa        &  53.8 $\pm$  4.1&  \textbf{55.7} $\pm$  3.6&  54.6 $\pm$  4.1&  54.1 $\pm$  4.8&  54.5 $\pm$  3.9\\
spambase    &  85.0 $\pm$  1.0&  86.2 $\pm$  1.1&  86.8 $\pm$  1.2&  88.1 $\pm$  1.1&  \textbf{88.7} $\pm$  0.8\\
wdbc        &  93.7 $\pm$  1.2&  93.2 $\pm$  2.1&  93.6 $\pm$  1.8&  92.9 $\pm$  1.6&  \textbf{94.7} $\pm$  1.4\\
iono        &  73.1 $\pm$  5.2&  74.8 $\pm$  3.3&  75.4 $\pm$  3.4&  73.3 $\pm$  4.1&  \textbf{75.6} $\pm$  3.4\\
pima        &  60.8 $\pm$  2.7&  60.9 $\pm$  2.4&  \textbf{62.2} $\pm$  1.7&  60.5 $\pm$  2.1&  60.5 $\pm$  1.9\\
wine        &  91.2 $\pm$  2.6&  92.7 $\pm$  3.8&  \textbf{94.2} $\pm$  3.8&  94.1 $\pm$  2.8&  93.5 $\pm$  3.1\\
glass       &  64.6 $\pm$  3.5&  \textbf{64.6} $\pm$  3.1&  61.1 $\pm$  4.9&  63.2 $\pm$  4.5&  64.2 $\pm$  5.1\\
newthyroid  &  86.6 $\pm$  4.6&  90.4 $\pm$  3.0&  91.1 $\pm$  4.9&  91.4 $\pm$  5.0&  \textbf{92.1} $\pm$  2.9\\
german      &  46.7 $\pm$  1.6&  \textbf{47.5} $\pm$  1.7&  47.3 $\pm$  2.3&  46.9 $\pm$  2.5&  46.4 $\pm$  1.8\\
vehicle     &  74.0 $\pm$  3.1&  85.5 $\pm$  3.1&  87.7 $\pm$  2.6&  \textbf{89.7} $\pm$  1.6&  88.5 $\pm$  2.6\\
spectfheart &  44.2 $\pm$  3.9&  45.5 $\pm$  5.4&  46.7 $\pm$  4.6&  42.6 $\pm$  8.0&  \textbf{49.2} $\pm$  5.0\\
hayes       &  63.4 $\pm$  9.0&  67.0 $\pm$  7.7&  64.7 $\pm$  6.7&  \textbf{67.7} $\pm$  7.4&  63.4 $\pm$  8.2\\
segmentation&  64.6 $\pm$  3.1&  64.3 $\pm$  2.9&  65.7 $\pm$  3.4&  70.4 $\pm$  2.4&  \textbf{74.6} $\pm$  1.9\\
abalone     &  \textbf{32.8} $\pm$  1.1&  31.6 $\pm$  1.0&  32.5 $\pm$  1.3&  32.5 $\pm$  1.4&  32.3 $\pm$  1.5\\
yeast       &  57.2 $\pm$  4.5&  59.7 $\pm$  3.8&  60.9 $\pm$  3.8&  60.8 $\pm$  4.6&  \textbf{61.8} $\pm$  3.0\\
libras      &  34.3 $\pm$ 10.6&  36.5 $\pm$ 12.6&  38.2 $\pm$ 12.2&  35.6 $\pm$ 10.9&  \textbf{41.5} $\pm$ 11.3\\
pageblocks  &  46.8 $\pm$  3.7&  48.3 $\pm$  4.5&  43.0 $\pm$  5.2&  \textbf{50.2} $\pm$  4.8&  49.1 $\pm$  4.1\\
\midrule
Mean        &  66.6 $\pm$  3.5&  68.2 $\pm$  3.5&  68.5 $\pm$  3.7&  69.1 $\pm$  3.8&  \textbf{70.2} $\pm$  3.3\\
\midrule
Average Rank & 4.04 & 3.30 & 2.65 & 2.96 & 2.04\\
\bottomrule
	\end{tabular}}
	\end{subtable}
\end{table*}

The results obtained are reported in Table \ref{tab:exp3_1} for SMOTE and in Table \ref{tab:exp3_2} for RUS.
Note that the results from Tables \ref{tab:exp2}, \ref{tab:exp3_1} and \ref{tab:exp3_2} were computed using the same training/test splits and the same validation folds and are thus comparable.
In each of the three settings considered, {\bf IML} obtains the best results showing that it is more appropriate for improving the F1-measure.
We also note that SMOTE allows one to increase significantly the performances of all  methods, while there is no gain with RUS in comparison with an approach without sampling.

This increase of performance suggests that SMOTE and {\bf IML} are more complementary than competitors with different objectives (re-balancing for the former and representation learning for the latter).

\paragraph{Third experiment---increasing the imbalance}
We now aim at showing the efficiency of our method by artificially increasing and decreasing the imbalance. % in the datasets.
For a given dataset, we create a maximum of $10$ synthetic variants where the percentage of minority examples is in \\$\{50\%, 40\%, 30\%, 20\%, 10\%, 5\%, 4\%, 3\%, 2\%, 1\%\}$.
To create a synthetic variant of a dataset with a percentage of minority examples higher than in the original dataset, we apply a random under sampling of the majority class until the desired percentage is reached.
Similarly, to create a synthetic variant with a smaller percentage of minority examples, we apply a random under sampling of the minority class. Note that we create the synthetic variant of the dataset only if it contains at least $20$ minority examples. 
For example, for the dataset {\sc spectfheart}, we cannot go under $10\%$ of minority examples. Due to the small number of minority examples present in the more imbalanced synthetic variants of the datasets, we split them into $50\%$ training and $50\%$ test examples.
We report the mean results over $20$ iterations where at each iteration we recompute the synthetic variants of the dataset and the train/test splits.
\begin{figure*}[t]
	\centering
	\includegraphics[width=1.00\textwidth]{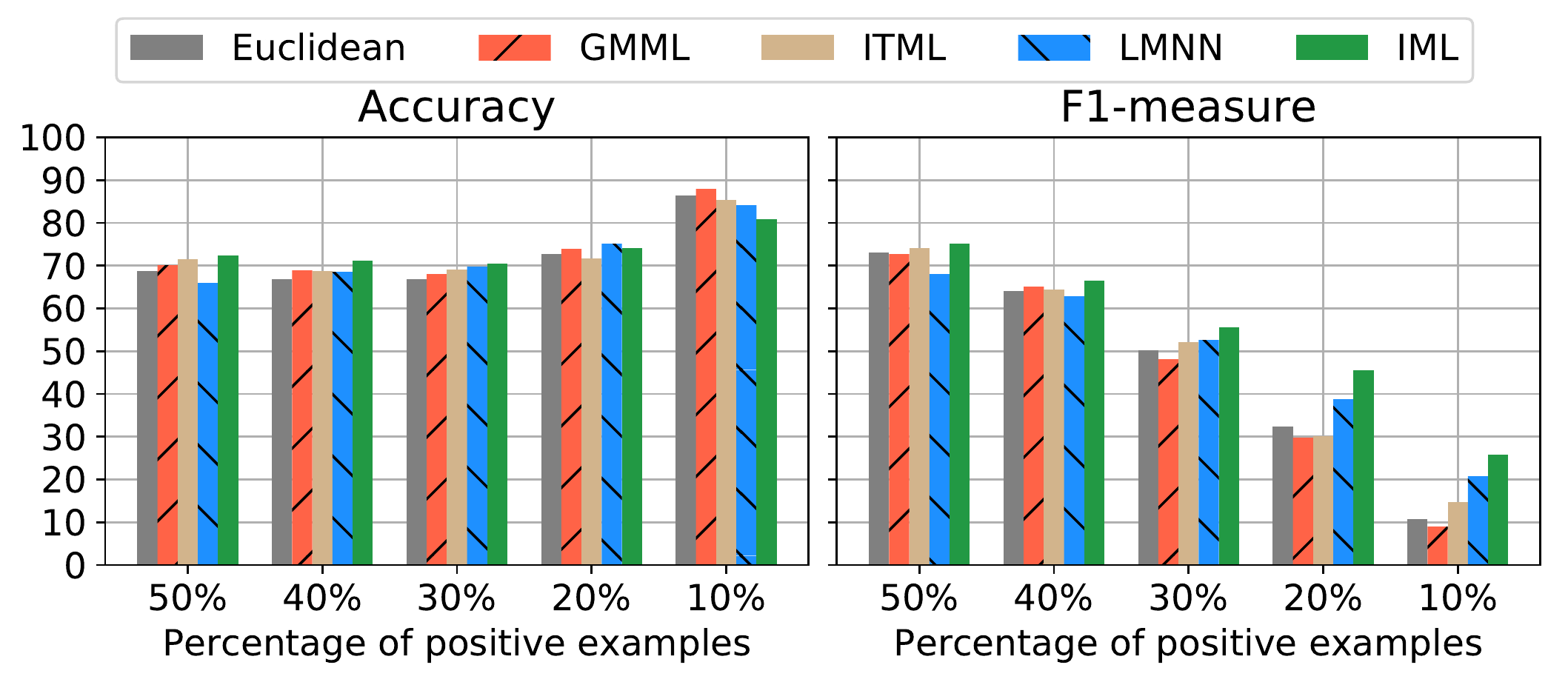}
	\caption{Mean Accuracy and F1-measure over $20$ splits on the {\sc spectfheart} dataset by artificially increasing the imbalance. We compare state-of-the-art metric learning algorithms with our proposed method {\bf IML}.}\label{fig:increaseImbalance2}
\end{figure*}

The results for the {\sc spectfheart} dataset (already used in the introduction of this paper) are reported in Figure \ref{fig:increaseImbalance2}. 
We see that like the other algorithms, {\bf IML} shows the same drop of F1-measure when increasing the imbalance, which shows the difficulty of learning from imbalanced data. 
However, it is important to notice that the drop of performances of {\bf IML} is the smallest among all algorithms.

\begin{figure*}[t]
	\centering
	\includegraphics[width=1.00\textwidth]{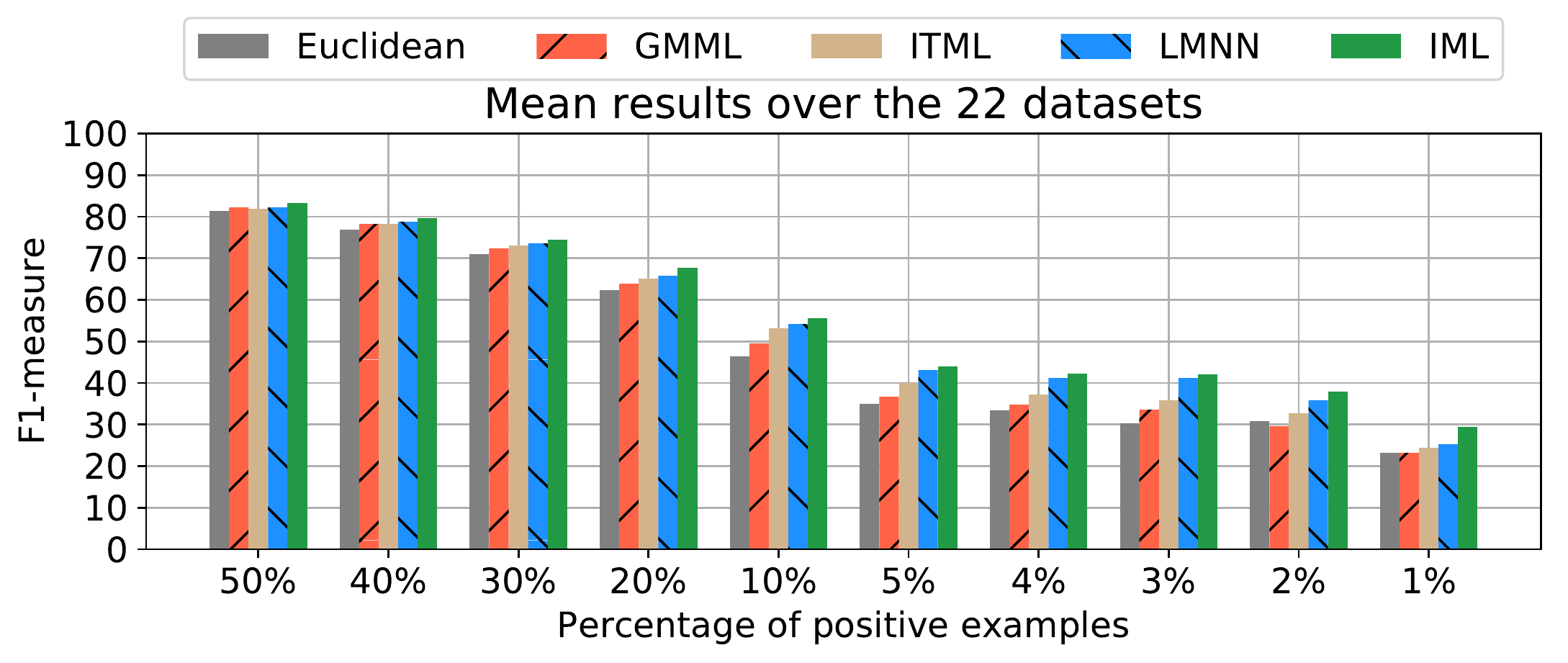}
	\caption{Mean F1-measure over $20$ splits and over the $22$ datasets by artificially increasing the imbalance. We compare state-of-the-art metric learning algorithms with our proposed method {\bf IML}.}\label{fig:increaseImbalance3}
\end{figure*}
To confirm the efficiency of {\bf IML} when facing imbalance data on a wide range of datasets, we present in Figure \ref{fig:increaseImbalance3} the results of the same experiments by averaging the results over all the datasets. 
We observe the same behavior as for the {\sc spectfheart} dataset.
Again, it is worth noticing that {\bf IML} is always more robust while facing imbalanced classes.

\paragraph{Fourth experiment---analyzing why {\bf IML} is better than the other metric learning algorithms on imbalanced data} 
When we described {\bf IML} in Section \ref{sec:iml}, we presented two strategies to deal with the imbalance. The first one, which is already used by some existing metric learning methods, is a selection of the similar and dissimilar pairs based on the nearest neighbor rule. The second one which we proposed in this paper is to weight differently the set of pairs based on the labels of the two examples composing the pairs. To see the impact of these two strategies, we compare in this last experiment {\bf IML} with two variants.

The variant called {\bf ML2} considers the loss of {\bf IML} but without the re-weighting of the set of pairs. 
Its loss is defined as follows:
\begin{align}
	\min_{\M\succeq 0}\ F(\M)\	 =\ &a\!\!\!\!\!\! \sum_{(z,z')\in \Simp}\!\!\!\!\!\!\,\ell_1(\M,z,z') +\nonumber\\
	&\ a\!\!\!\!\!\!\sum_{(z,z')\in \Simn}\!\!\!\!\!\!\!\!\,\ell_1(\M,z,z') +\nonumber\\
	&\ (1-a)\!\!\!\!\!\!\sum_{(z,z')\in \Disp}\!\!\!\!\!\!\,\ell_2(\M,z,z') +\nonumber\\
	&\ (1-a)\!\!\!\!\!\!\sum_{(z,z')\in \Disn}\!\!\!\!\!\!\,\ell_2(\M,z,z') +\nonumber\\
	&\  \lambda \left\Vert\M-\Id\right\Vert^2_F,\,\label{eq:problemML2}
\end{align}
where the difference with \eqref{eq:problem} is that we no longer multiply each of the four sets by $\tfrac{1}{4\vert \text{set}\vert}$.

The variant called {\bf ML1} considers the same loss as {\bf ML2}, but we select randomly the pairs of examples. 
In order to use the same number of pairs in {\bf ML1} and {\bf IML}, we draw randomly $2nk$ pairs for {\bf ML1} since {\bf IML} considers $k$ similar pairs and $k$ dissimilar pairs per training example.

\begin{figure*}[t]
	\centering
	\includegraphics[width=1.00\textwidth]{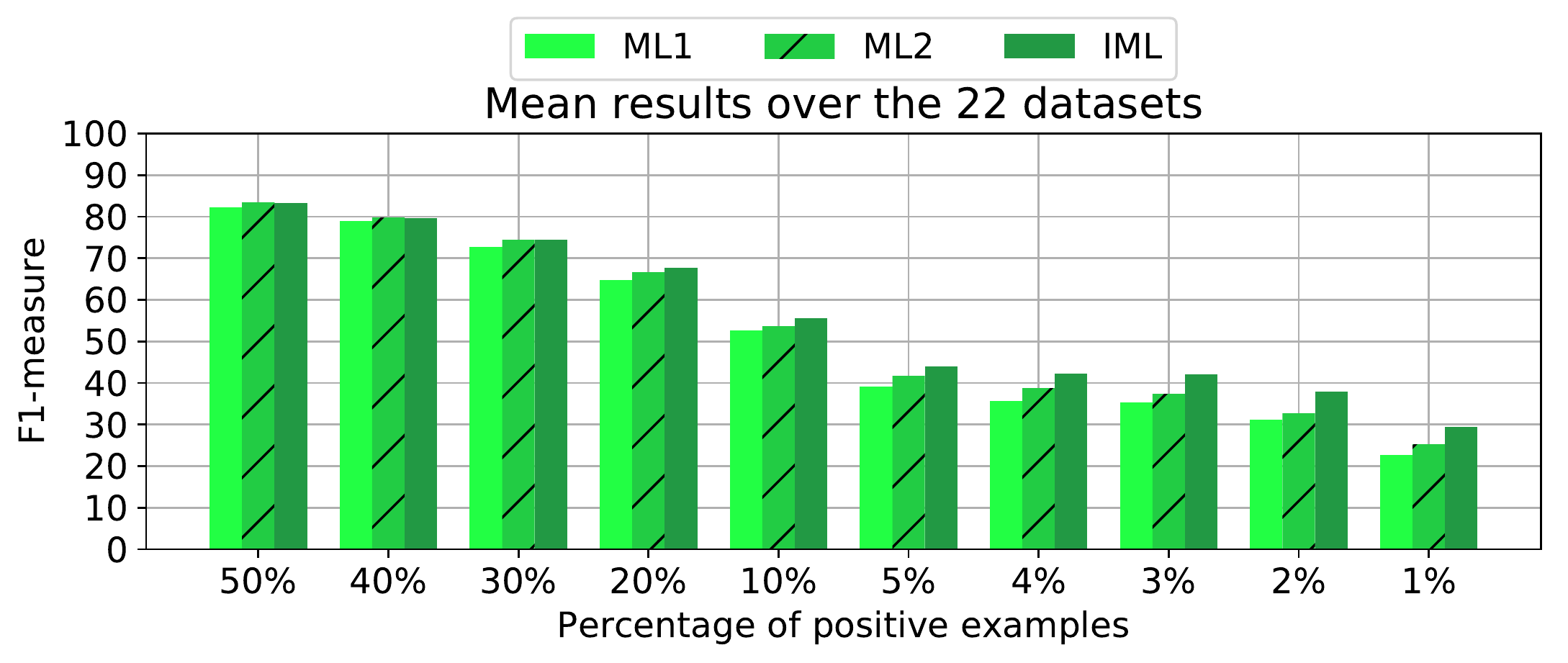}
	\caption{Mean F1-measure over $20$ splits and over the $22$ datasets by artificially increasing the imbalance. We compare two variants of {\bf IML}. The variant {\bf ML2} removes one component that allows {\bf IML} to deal with the imbalance: the re-weighting of the pairs. {\bf ML1} removes in addition another component to deal with the imbalance: instead of selecting the similar and dissimilar pairs with the nearest neighbor rule, they are selected randomly by {\bf ML1}.}\label{fig:increaseImbalance4}
\end{figure*}

The results of this experiment are reported in Figure~\ref{fig:increaseImbalance4}. 
When the classes are balanced with $50\%$ of minority examples, we observe that {\bf IML} and its two variants present the same performances. 
When increasing the imbalance, as expected {\bf IML} tends to be better than {\bf ML2} itself better than {\bf ML1}. 
This shows that our two strategies to deal with imbalanced data do not degrade the results on balanced data and that they are complementary to improve the performances in an imbalanced setting.

\section{Conclusion and perspectives}
\label{sec:conclusion}
In this paper, we revisit the classic formulation of metric learning algorithms that learn a Mahalanobis metric in the light of imbalanced data issues.
Our method resorts to two complementary strategies to deal with the imbalance.
First, unlike the state of the art methods that do not make any distinction between the pairs, we propose to decompose the usual loss with respect to the different possible labels involved in the pairs of examples.
This decomposition allows us to assign specific weights to each type of pairs in order to improve the performance on the minority class.
Second, contrarily to many metric algorithms that select the pairs of similar and dissimilar examples randomly, we select them based on the nearest neighbor rule.
Our experimental evaluation shows that we are able to obtain better results than state of the art metric learning algorithms in terms of \mbox{F1-measure} over balanced and imbalanced datasets.
Last but not least, artificially increasing the imbalance in the datasets shows that our two strategies to deal with the imbalance are complementary.

We believe that our work gives rise to exciting perspectives when facing imbalanced data.
Among them, we want to study how our algorithm could be adapted to learn non-linear metrics.
From an algorithmic point of view, we would like to extend our method by deriving a closed form solution in a similar way as done by Zadeh {\it et al.}~\cite{zadeh2016geometric} to drastically reduce the computation time while maintaining good performances. 
Finally, Problem~\eqref{eq:problem} opens the door to the derivation of generalization guarantees based for example on the uniform stability framework.

\bibliographystyle{plain}
\bibliography{arxiv}

\begin{thebibliography}{10}

\bibitem{Aggarwal}
C.~Aggarwal.
\newblock {\em Outlier Analysis}.
\newblock Springer, 2013.

\bibitem{bauder2018data}
R.~Bauder, T.~Khoshgoftaar, and T.~Hasanin.
\newblock Data sampling approaches with severely imbalanced big data for
  medicare fraud detection.
\newblock In {\em ICTAI}, pages 137--142. IEEE, 2018.

\bibitem{bellet2013survey}
A.~Bellet, A.~Habrard, and M.~Sebban.
\newblock A survey on metric learning for feature vectors and structured data.
\newblock {\em arXiv preprint arXiv:1306.6709}, 2013.

\bibitem{bellet2015metric}
A.~Bellet, A.~Habrard, and M.~Sebban.
\newblock Metric learning.
\newblock {\em Synthesis Lectures on Artificial Intelligence and Machine
  Learning}, 9(1), 2015.

\bibitem{branco2016survey}
P.~Branco, L.~Torgo, and R.~Ribeiro.
\newblock A survey of predictive modeling on imbalanced domains.
\newblock {\em ACM Computing Surveys (CSUR)}, 49(2):31, 2016.

\bibitem{cao2016generalization}
Q.~Cao, Z.~Guo, and Y.~Ying.
\newblock Generalization bounds for metric and similarity learning.
\newblock {\em Machine Learning}, 102(1):115--132, 2016.

\bibitem{chandola2009anomaly}
V.~Chandola, A.~Banerjee, and V.~Kumar.
\newblock Anomaly detection: A survey.
\newblock {\em ACM computing surveys (CSUR)}, 41(3):15, 2009.

\bibitem{Chawla:2002}
N.~Chawla, K.~Bowyer, L.~Hall, and P.~Kegelmeyer.
\newblock Smote: Synthetic minority over-sampling technique.
\newblock {\em Journal of Artificial Intelligence Research}, 16(1):321--357,
  June 2002.

\bibitem{chawla2003smoteboost}
N.~Chawla, A.~Lazarevic, L.~Hall, and K.~Bowyer.
\newblock Smoteboost: Improving prediction of the minority class in boosting.
\newblock In {\em PKDD}, pages 107--119. Springer, 2003.

\bibitem{davis2007information}
J.~Davis, B.~Kulis, P.~Jain, S.~Sra, and I.~Dhillon.
\newblock Information-theoretic metric learning.
\newblock In {\em ICML}, 2007.

\bibitem{douzas2018effective}
G.~Douzas and F.~Bacao.
\newblock Effective data generation for imbalanced learning using conditional
  generative adversarial networks.
\newblock {\em Expert Systems with applications}, 91:464--471, 2018.

\bibitem{drummond2003c4}
C.~Drummond and R.~Holte.
\newblock C4. 5, class imbalance, and cost sensitivity: why under-sampling
  beats over-sampling.
\newblock In {\em Workshop on learning from imbalanced datasets II}, volume~11,
  pages 1--8. Citeseer, 2003.

\bibitem{elkan2001foundations}
C.~Elkan.
\newblock The foundations of cost-sensitive learning.
\newblock In {\em IJCAI}, pages 973--978, 2001.

\bibitem{estabrooks2004multiple}
A.~Estabrooks, T.~Jo, and N.~Japkowicz.
\newblock A multiple resampling method for learning from imbalanced data sets.
\newblock {\em Computational intelligence}, 20(1):18--36, 2004.

\bibitem{feng2018learning}
L.~Feng, H.~Wang, B.~Jin, H.~Li, M.~Xue, and L.~Wang.
\newblock Learning a distance metric by balancing kl-divergence for imbalanced
  datasets.
\newblock {\em IEEE Transactions on Systems, Man, and Cybernetics: Systems},
  2018.

\bibitem{ferreira2017improving}
L.~Ferreira, J.~Barddal, F.~Enembreck, and H.~Gomes.
\newblock Improving credit risk prediction in online peer-to-peer (p2p) lending
  using imbalanced learning techniques.
\newblock In {\em ICTAI}, pages 175--181. IEEE, 2017.

\bibitem{FreryECML2017}
J.~Frery, A.~Habrard, M.~Sebban, O.~Caelen, and L.~He-Guelton.
\newblock Efficient top rank optimization with gradient boosting for supervised
  anomaly detection.
\newblock In {\em ECML-PKDD}, 2017.

\bibitem{galar2012review}
M.~Galar, A.~Fernandez, E.~Barrenechea, H.~Bustince, and F.~Herrera.
\newblock A review on ensembles for the class imbalance problem: bagging-,
  boosting-, and hybrid-based approaches.
\newblock {\em IEEE Transactions on Systems, Man, and Cybernetics},
  42(4):463--484, 2012.

\bibitem{han2005borderline}
H.~Han, W.~Wang, and B.~Mao.
\newblock Borderline-smote: a new over-sampling method in imbalanced data sets
  learning.
\newblock In {\em ICIC}, pages 878--887. Springer, 2005.

\bibitem{he2009learning}
H.~He and E.~Garcia.
\newblock Learning from imbalanced data.
\newblock {\em IEEE TKDE}, 21(9), 2009.

\bibitem{jin2009regularized}
R.~Jin, S.~Wang, and Y.~Zhou.
\newblock Regularized distance metric learning: Theory and algorithm.
\newblock In {\em NIPS}, 2009.

\bibitem{kulis2013metric}
B.~Kulis.
\newblock Metric learning: A survey.
\newblock {\em Foundations and Trends in Machine Learning}, 5(4):287--364,
  2013.

\bibitem{lee2008rank}
J.~Lee, R.~Jin, and A.~Jain.
\newblock Rank-based distance metric learning: An application to image
  retrieval.
\newblock In {\em CVPR}, 2008.

\bibitem{lemaitre2017imbalanced}
G.~Lema{\^\i}tre, F.~Nogueira, and C.~Aridas.
\newblock Imbalanced-learn: A python toolbox to tackle the curse of imbalanced
  datasets in machine learning.
\newblock {\em The Journal of Machine Learning Research}, 18(1):559--563, 2017.

\bibitem{liu2009exploratory}
X.~Liu, J.~Wu, and Z.~Zhou.
\newblock Exploratory undersampling for class-imbalance learning.
\newblock {\em IEEE Transactions on Systems, Man, and Cybernetics, Part B
  (Cybernetics)}, 39(2):539--550, 2009.

\bibitem{lopez2013insight}
V.~L{\'o}pez, A.~Fern{\'a}ndez, S.~Garc{\'\i}a, V.~Palade, and F.~Herrera.
\newblock An insight into classification with imbalanced data: Empirical
  results and current trends on using data intrinsic characteristics.
\newblock {\em Information sciences}, 250:113--141, 2013.

\bibitem{lu2014neighborhood}
J.~Lu, X.~Zhou, Y.~Tan, Y.~Shang, and J.~Zhou.
\newblock Neighborhood repulsed metric learning for kinship verification.
\newblock {\em IEEE Transactions on Pattern Analysis and Machine Intelligence},
  36(2):331--345, 2014.

\bibitem{mcfee2010metric}
B.~McFee and G.~Lanckriet.
\newblock Metric learning to rank.
\newblock In {\em ICML}, 2010.

\bibitem{pereira2018dealing}
R.~Pereira, Y.~Costa, and C.~Silla.
\newblock Dealing with imbalanceness in hierarchical multi-label datasets using
  multi-label resampling techniques.
\newblock In {\em ICTAI}, pages 818--824. IEEE, 2018.

\bibitem{scholkopfBV95}
B.~Sch{\"o}lkopf, C.~Burges, and V.~Vapnik.
\newblock Extracting support data for a given task.
\newblock In {\em KDD}, pages 252--257, 1995.

\bibitem{schultz2004learning}
M.~Schultz and T.~Joachims.
\newblock Learning a distance metric from relative comparisons.
\newblock In {\em NIPS}, 2004.

\bibitem{fmesure}
C.~van Rijsbergen.
\newblock Further experiments with hierarchic clustering in document retrieval.
\newblock {\em Information Storage and Retrieval}, 10(1):1 -- 14, 1974.

\bibitem{vapnik1995nature}
V.~Vapnik.
\newblock The nature of statistical learning theory.
\newblock 1995.

\bibitem{vogel2018probabilistic}
R.~Vogel, A.~Bellet, and S.~Cl{\'e}men{\c{c}}on.
\newblock A probabilistic theory of supervised similarity learning for
  pointwise {ROC} curve optimization.
\newblock {\em ICML}, 2018.

\bibitem{wang2018iterative}
N.~Wang, X.~Zhao, Y.~Jiang, and Y.~Gao.
\newblock Iterative metric learning for imbalance data classification.
\newblock In {\em IJCAI}, 2018.

\bibitem{weinberger2008fast}
K.~Weinberger and L.~Saul.
\newblock Fast solvers and efficient implementations for distance metric
  learning.
\newblock In {\em ICML}, pages 1160--1167. ACM, 2008.

\bibitem{weinberger2009distance}
K.~Weinberger and L.~Saul.
\newblock Distance metric learning for large margin nearest neighbor
  classification.
\newblock {\em JMLR}, 10(Feb):207--244, 2009.

\bibitem{xiang2008learning}
S.~Xiang, F.~Nie, and C.~Zhang.
\newblock Learning a mahalanobis distance metric for data clustering and
  classification.
\newblock {\em Pattern Recognition}, 41(12):3600--3612, 2008.

\bibitem{xing2003distance}
E.~Xing, M.~Jordan, S.~Russell, and A.~Ng.
\newblock Distance metric learning with application to clustering with
  side-information.
\newblock In {\em NIPS}, 2003.

\bibitem{zadeh2016geometric}
P.~Zadeh, R.~Hosseini, and S.~Sra.
\newblock Geometric mean metric learning.
\newblock In {\em ICML}, 2016.

\bibitem{zadrozny2003cost}
B.~Zadrozny, J.~Langford, and N.~Abe.
\newblock Cost-sensitive learning by cost-proportionate example weighting.
\newblock In {\em ICDM}. IEEE, 2003.

\bibitem{zheng2011person}
W.~Zheng, S.~Gong, and T.~Xiang.
\newblock Person re-identification by probabilistic relative distance
  comparison.
\newblock In {\em CVPR 2011}, pages 649--656. IEEE, 2011.

\bibitem{zhu1997algorithm}
C.~Zhu, R.~Byrd, P.~Lu, and J.~Nocedal.
\newblock Algorithm 778: L-bfgs-b: Fortran subroutines for large-scale
  bound-constrained optimization.
\newblock {\em ACM TOMS}, 23(4):550--560, 1997.

\end{thebibliography}

\end{document}